# Speed-reading Tool Powered by Artificial Intelligence for Students with ADHD, Dyslexia, and Short Attention Span

Megat Irfan Zackry Bin Ismail
*Software Engineering*
*Universiti Teknologi Malaysia*
*Johor Bahru, Malaysia*

Ahmad Nazran Bin Yusri
*Software Engineering*
*Universiti Teknologi Malaysia*
*Johor Bahru, Malaysia*

Muhammad Hafizzul Bin Abdul Manap
*Software Engineering*
*Universiti Teknologi Malaysia*
*Johor Bahru, Malaysia*

Muhammad Muizzuddin Bin Kamarozaman
*Software Engineering*
*Universiti Teknologi Malaysia*
*Johor Bahru, Malaysia*

***Abstract.*** *This paper presents an artificial intelligence tool designed to assist students with dyslexia, ADHD, and short attention spans in processing text-based information more efficiently. The proposed solution addresses both cognitive and visual reading barriers by pairing a cloud-hosted large language model with adaptive typographic formatting. At its core, the tool streams a request to Google's Gemini API to generate accurate, context-aware summaries in real time, removing the need to host or fine-tune a local model. The application accepts pasted text as well as uploaded .txt, .pdf, and .docx files, extracting and summarizing each source independently. To further optimize readability, the system layers a custom half-word bolding technique with a second pass that identifies and highlights genuine keywords using part-of-speech tagging and frequency ranking, and lets users manually adjust line, word, letter, and text spacing. Deployed through a lightweight Flask web framework that proxies requests to Gemini, the system provides a highly personalized and accessible user interface. Initial results demonstrate that this integrated approach significantly improves both reading speed and comprehension, enabling targeted students to digest complex textual data with greater focus.*

## I. INTRODUCTION

The advent of the digital age has caused an explosion of information, making it difficult for many individuals to process vast amounts of daily text. This challenge is notably amplified for adolescents and students diagnosed with specific reading disabilities and attention deficit hyperactivity disorder [1]. To address these cognitive and visual reading barriers, this paper proposes an artificial intelligence speed-reading tool. Rather than hosting and fine-tuning a summarization model locally, the application calls Google's Gemini large language model through its Interactions API and streams the resulting summary back to the reader as it is produced. The tool also accepts a mix of pasted text and uploaded files, including plain text, PDF, and Word documents, so a student can summarize several assigned readings in a single session. Together with adjustable typography, this enables more efficient and accessible information consumption across educational and research domains.

## II. BACKGROUND

Traditional speed-reading techniques often require significant training and practice, which can be exhausting for neurodivergent learners. Recent advances in deep learning have introduced automated text-processing and comprehension pipelines built on the Transformer architecture, which relies on a self-attention mechanism to weigh the importance of different words in a sequence [2]. This parallel processing capability allows the model to efficiently handle long texts compared to traditional sequential algorithms, and feedforward neural networks within the architecture let it learn complex text patterns without losing contextual accuracy.

The proposed implementation instead calls Gemini [3], a large, cloud-hosted Transformer-based model, through Google's Interactions API. This shifts the computational burden off the client device, removes the need to manage model weights or a GPU, and gives the tool access to a model trained at a scale a small fine-tuning run could not match.

## III. TEXT SUMMARIZATION METHODOLOGY

The summarization engine now calls Google's Gemini Interactions API [4] rather than running a locally hosted model. Text originating from the browser, whether pasted directly or extracted from an uploaded file, is sent to a small Flask backend, which forwards it to Gemini together with a fixed instruction prompt asking for a clear, concise summary that preserves key facts, names, numbers, and conclusions without added commentary.

The backend exists for two reasons. First, the Interactions API cannot currently be called directly from browser JavaScript because of a cross-origin restriction on one of its request headers. Second, routing the call through a server keeps the user's API key out of client-side network requests, consistent with Google's own guidance. Responses are requested in streaming mode, so the backend relays the incoming server-sent events to the browser as they arrive, and the summary appears on screen as it is generated rather than only once the full response completes.

Multiple sources can be summarized at once. Pasted text and every uploaded file are treated as independent sources; each streamed concurrently with its own status and a dropdown that lets the reader switch between them.

File ingestion happens before any call to Gemini. Plain text files are decoded with a UTF-8, UTF-16, then Latin-1 fallback. PDFs are parsed, with a best-effort attempt to open documents protected with an empty password, and pages that yield no extractable text, such as scanned image-only pages, are skipped. Word documents are parsed with python-docx, pulling text from both paragraphs and tables. A file that fails to yield usable text, whether it is empty, an image-only PDF, or an unsupported format such as legacy .doc, returns a clear per-file error instead of failing the whole batch.

## IV. TYPOGRAPHIC ENHANCEMENTS

Beyond summarization, the tool applies a hybrid formatting pass to the finished summary. As the summary streams in, the browser immediately renders a live half-word bolded version so the reader has something to follow right away. Once the stream finishes, a second request recomputes the same half-word bolding against the complete text and layers real keyword highlighting on top of it.

Rather than bolding or highlighting every word indiscriminately, this second pass identifies genuine keywords and key phrases. Highlighting key content in this more selective, visual manner has been shown to improve reading speed and text comprehension for various types of learners [5]. The finished summary is tagged for part of speech using NLTK; nouns, proper nouns, adjectives, and numbers are treated as candidate keywords, common stop words are filtered out, and the remaining candidates are ranked by frequency, with proper nouns and numbers given a small boost since a single mention of a name or statistic is often still worth surfacing. A fixed proportion of the top-ranked candidates are highlighted directly in the summary text and also listed as standalone keyword chips beneath it, giving the reader both an inline and an at-a-glance view of the material's key terms.

Readers can also manually adjust line spacing, word spacing, letter spacing, and text size to suit their personal visual processing preferences, and the interface reports an estimated reading time alongside each summary. These dynamic features are delivered through the same lightweight Flask web framework, which provides a highly responsive and user-friendly interface built with standard web technologies.

## V. RESULT & DISCUSSION

Initial deployment of the application demonstrated a substantial improvement in reading efficiency among the test group of students, particularly those with severe short attention spans. Condensing large volumes of text into succinct summaries has proven highly effective for processing complex information [6]. Moving summarization to a hosted large language model removed the overhead of shipping and running model weights locally, while the addition of file uploads and the new keyword-highlighting pass let the same summary serve two reading strategies at once: a quick scan of the highlighted terms, or a full read of the bolded text. The combination of a large language model, adjustable typography, and multi-format file input creates a highly accessible reading environment that mitigates common cognitive barriers.

## VI. FIGURES

The figure below is a screenshot of the user interface that acts as the medium through which the user summarizes a piece of text.

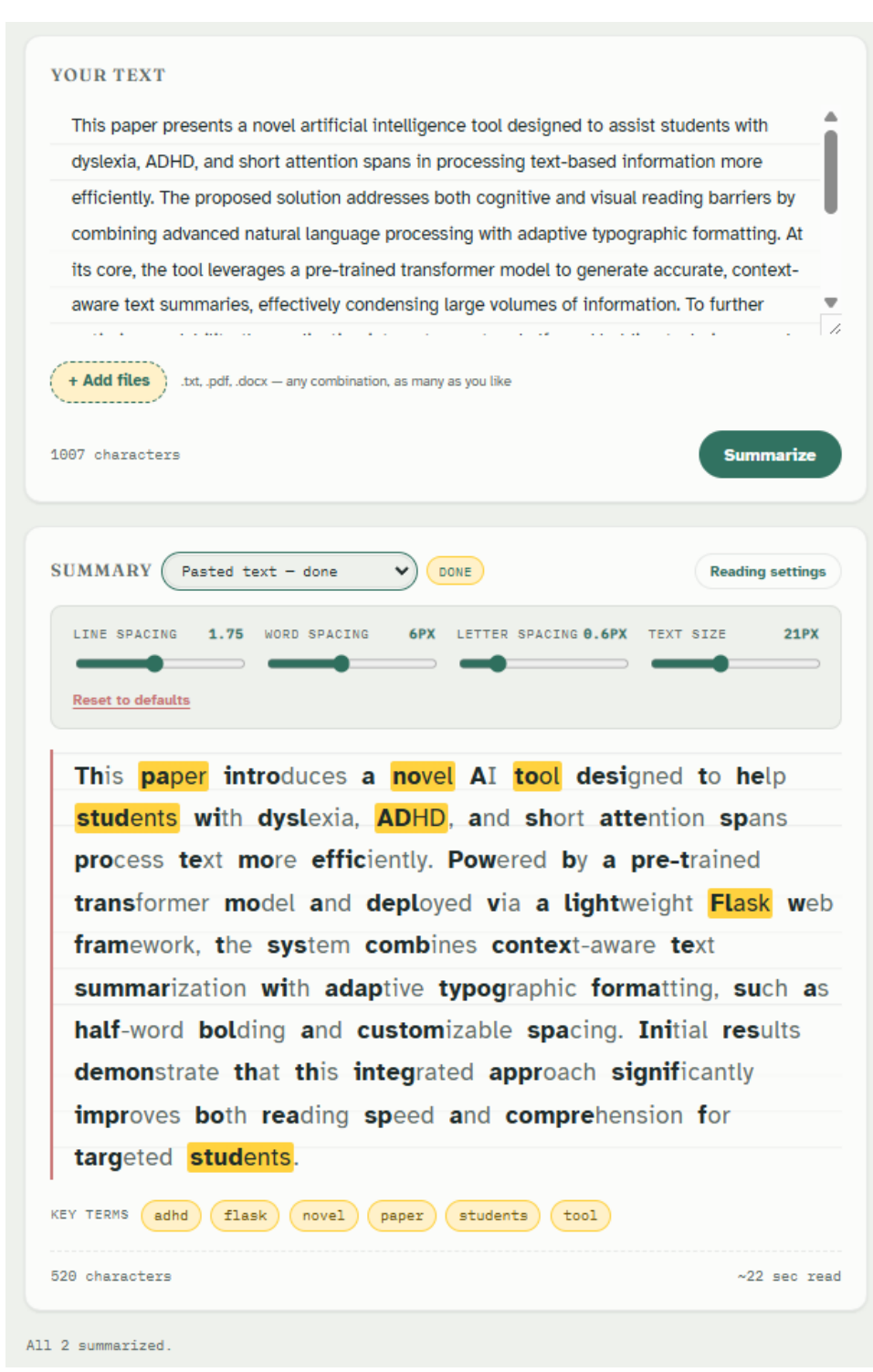

*Fig. 1. Summarizer User Interface*

## VII. CONCLUSION

The AI-based speed-reading tool proposed in this project has shown promising results in improving reading speed and comprehension among students with dyslexia, ADHD, and especially those with severe short attention span. Moving the summarization engine to Google's Gemini API and adding support for pasted text alongside .txt, .pdf, and .docx uploads has made the tool more capable without adding local infrastructure to maintain, and the new keyword-highlighting layer gives readers a second, more selective way to scan a summary on top of the original half-word bolding. It's important to note that the effectiveness of the tool may vary depending on the complexity of the text and the individual's reading habits, and that summary quality is now also dependent on the upstream Gemini model and API availability.

Future work will focus on refining the prompt and file-ingestion pipeline to better handle a wider range of text complexities, and on exploring additional features such as more sophisticated ranking for the keyword-highlighting pass, OCR support for scanned PDFs, and further customization options for spacing adjustments. User feedback will continue to be an invaluable resource for guiding future improvements and ensuring the tool remains responsive to the needs of its users.

## REFERENCES

[1] K. Ghelani, R. Sidhu, U. Jain, and R. Tannock, "Reading comprehension and reading related abilities in adolescents with reading disabilities and attention-deficit/hyperactivity disorder," Dyslexia, vol. 10, no. 4, pp. 364-384, Nov. 2004, doi: 10.1002/DYS.285.

[2] A. Vaswani et al., "Attention Is All You Need," p. 1, Jun. 2017, Accessed: Jul. 24, 2026. [Online]. Available: https://arxiv.org/pdf/1706.03762

[4] Gemini Team, Google, "Gemini: A Family of Highly Capable Multimodal Models," arXiv:2312.11805, Dec. 2023, Accessed: Jul. 25, 2026. [Online]. Available: https://arxiv.org/abs/2312.11805

[5] Google, "Interactions API quickstart," Gemini API Documentation, Accessed: Jul. 25, 2026. [Online]. Available: https://ai.google.dev/gemini-api/docs/interactions/quickstart

[6] M. Khost AlJarkas and M. K. AlJarkas, "The Influence of Bolded Keywords in Online News Texts on Reading Speed and Comprehension for Heritage+ Learners of Arabic," Theses and Dissertations, Jun. 2023, Accessed: Jul. 24, 2026. [Online]. Available: https://fount.aucegypt.edu/etds/2145

[7] P. Singh, P. Kashetty, A. S. R. T. Reddy, G. S. Teja, and V. Anusha, "Automated News Summarization using Transformers," ISML 2024 - Intelligent Systems and Machine Learning Conference, pp. 693-697, 2024, doi: 10.1109/ISML60050.2024.11007399.